%% file: main.tex
\documentclass[10pt,twocolumn,letterpaper]{article}

\usepackage{cvpr}
\input{preamble}

\definecolor{cvprblue}{rgb}{0.21,0.49,0.74}
\usepackage[pagebackref,breaklinks,colorlinks,allcolors=cvprblue]{hyperref}
\usepackage{xcolor}         
\usepackage{graphicx}
\usepackage{booktabs}
\usepackage{multirow}
\usepackage{multicol}
\usepackage{bm}
\usepackage{algorithm} %
\usepackage{amssymb}
\usepackage{amsmath}
\usepackage{amsmath}
\usepackage{color}
\usepackage{xcolor}
\usepackage{colortbl}
\usepackage{circledsteps}
\usepackage{algpseudocode}
\usepackage{makecell}
\definecolor{lightgray}{gray}{0.9}

\def\paperID{ } 
\def\confName{CVPR}
\def\confYear{2026}

\title{\mtd{}: Generating LiDAR World Model via Deformable Mamba}

\author{Yang Wu$^{1}$ \quad Zhaojiang Liu$^3$ \quad Qiang Meng$^{\dagger}$ \quad Youquan Liu$^4$ \quad \\ 
Renliang Weng$^5$ \quad Jianjun Qian$^1$ \quad Jian Yang$^{1,2}$ \quad Jin Xie$^{2}$\thanks{Corresponding author. $^{\dagger}$ Independent researcher.} \\
\vspace{-0.2cm}
$^1$PCA Lab@NJUST \quad $^2$NJU \quad  $^3$SJTU \quad  $^4$FDU \quad $^5$NTU  \\
{\tt\small wuyang98@njust.edu.cn; csjxie@nju.edu.cn} 
}

\begin{document}
\maketitle

\input{sec/0_abstract}
\input{sec/1_intro}
\input{sec/2_related}
\input{sec/3_method}
\input{sec/4_exp}
\input{sec/5_conclu}

\clearpage
\clearpage
\textbf{Acknowledgments.} This work was supported by the National Key R\&D Program of China No. 2024YFC3015801, NSFC under Grant Nos. U24A20330, 62361166670, 62276144, and Basic Research Program of Jiangsu under Grant No. BK20253028.
\vspace{-0.27cm}

{
    \small
    \bibliographystyle{ieeenat_fullname}
    \bibliography{main}
}


\end{document}

%% file: preamble.tex
\def\mtd{GEM}
\def\moduleone{LiDAR scene tokenizer}
\def\moduletwo{dynamic-static separator}
\def\modulethree{tri-path deformable Mamba}
\def\modulefour{adaptive gated attention}

\usepackage[
skip=7pt]{caption}

%% file: sec/0_abstract.tex
\vspace{-0.27cm}
\begin{abstract}
    World models, which simulate environmental dynamics and generate sensor observations, are gaining increasing attention in autonomous driving. 
    However, progress in LiDAR-based world models has lagged behind those built on camera videos or occupancy data, primarily due to two core challenges: the inherent disorder of LiDAR point clouds and the difficulty of distinguishing dynamic objects from static structures.
    To address these issues, we propose \textbf{\mtd{}}: a \textbf{G}enerative LiDAR world model that leverages d\textbf{E}formable \textbf{M}amba architecture, significantly improving fidelity and imaginative capability.
    Specifically, leveraging the structural similarity between sequential laser scanning and Mamba's processing mechanism, we first tokenize LiDAR sweeps into compact representations via a custom \moduleone{}.
    After unsupervised disentanglement of tokenized features via a \moduletwo{}, a \modulethree{} is introduced to perform selective scanning and adaptive gating fusion over the disentangled features, leading to enhanced spatial-temporal understanding of the world evolution.
    Optionally, a planner and a BEV layout controller can be integrated to explore the model's capability for autonomous rollout and its potential to generate ``what-if" scenarios.
    Extensive experiments show that \mtd{} achieves state-of-the-art performances across diverse benchmarks and evaluation settings, demonstrating its superiority and effectiveness. Project page: \url{https://github.com/wuyang98/GEM}.
\end{abstract}
\vspace{-0.27cm}

%% file: sec/1_intro.tex
\vspace{-0.27cm}
\section{Introduction}\label{sec:intro}
\begin{figure}[!t]
\includegraphics[width=0.478\textwidth]{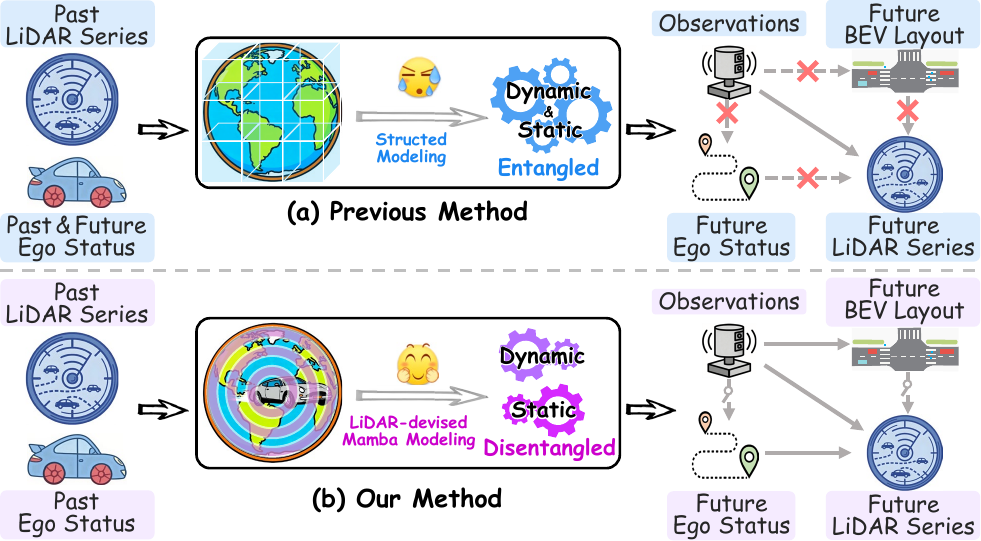}
\caption{(a) Previous methods~\cite{weng2021inverting,weng2022s2net,liang2025learning,khurana2023point,zhang2023copilot4d} fail to adequately model LiDAR-specific properties, suffer from entangled dynamic-static features, and must rely on ground truth ego status, limiting their ability to understand the driving world.
(b) In contrast, our approach explicitly disentangles dynamic and static elements following LiDAR's physical structure.
This enables more accurate scene comprehension while supporting autonomous rollout and reliable ``what-if" reasoning.
}
\vspace{-0.27cm}
\label{fig: intro}
\end{figure}

World models represent a transformative technology, enabling autonomous vehicles to evolve from passively reacting to their environment to actively reasoning about the future~\cite{kong20253d}. 
While significant progress has been made for camera video-based and occupancy-based  methods~\cite{zheng2024occworld, alhaija2025cosmos}, the potential of LiDAR-based world models remains largely unexplored, despite LiDAR's inherent advantage in providing precise geometric capture of driving environments.

Developing a robust LiDAR world model faces two primary challenges. 
The first is the unstructured nature of point clouds, which prevents the direct application of well-developed techniques for structured data like camera video and occupancy.
This fundamental mismatch compels existing methods to efficiently convert point clouds into intermediate representations. 
For instance, some methods~\cite{khurana2023point, zhang2023copilot4d} project LiDAR data into dense voxel or Bird's-Eye-View (BEV) features. However, these conversions sacrifice fine-grained geometric detail, consequently degrading prediction fidelity.
Other approaches~\cite{weng2021inverting,weng2022s2net,qu2025self,liang2025learning} instead map point clouds into range images to leverage conventional feature extractors.
A key limitation of these approaches is their reliance on convolution- or transformer-based architectures, which remain inherently misaligned with the sequential scanning mechanism of laser sweeps. 

The second challenge stems from the inherent ambiguity in raw point clouds, which lack discriminative cues like texture from camera videos or semantic labels from occupancy data.
This absence hinders LiDAR world models from effectively comprehending the evolving patterns of dynamic objects from the static environment. 
As shown in \cref{fig: intro}(a), previous methods~\cite{weng2021inverting,weng2022s2net,liang2025learning,khurana2023point,zhang2023copilot4d} fail to provide explicit scene decomposition, resulting in reduced geometric accuracy, noticeable temporal inconsistencies, and limited environmental comprehension.
Furthermore, these approaches lack autonomous rollout and controllable generation capabilities, which are critical for scalable world modeling.

In this work, we propose a LiDAR world model named \textbf{\mtd{}} to address the aforementioned challenges.
As illustrated in \cref{fig: intro} (b), \mtd{} is built on three key innovations:
(1) a compact representation dedicated to LiDAR scanning; 
(2) the explicit disentanglement of dynamic and static elements;
and (3) optional support for autonomous rollout (autonomously planning future ego status and forecasting future observations) and controllable generation. 

Our approach is grounded in the structural similarity between sequential laser scanning and the processing mechanism of Mamba~\cite{gu2023mamba}.
Inspired by this, we first introduce a \moduleone{} on top of the Mamba architecture, which excels at preserving the inherent physical structure of LiDAR data for robust sequential modeling.
Building upon this representation, we explicitly disentangle the dynamic and static scene components to enhance the model's comprehension of world evolution. 
Specifically, we devise an unsupervised \moduletwo{} to separate dynamic and static features without costly annotations. 
The disentangled features are then processed by a \modulethree{} that captures both local object evolution and global scene context in parallel. 
Finally, an \modulefour{} dynamically evaluates the importance of dynamic and static features and fuses them into a coherent scene representation.
On nuScenes~\cite{caesar2020nuScenes} and KITTI Odometry~\cite{geiger2012we}, \mtd{} achieves new state-of-the-art performance for 1s and 3s predictions. 
Beyond the core world modeling, \mtd{} can integrate autonomous rollout and external guidance, 
enabling controllable ``what-if" scenario reasoning for driving world.
Contributions are summarized as follows: 
\begin{itemize}
\item We introduce a \moduleone{} based on the Mamba architecture, designed for compact and structurally faithful latent representation of LiDAR data.
\item We present a LiDAR world model with explicit dynamic-static disentanglement in driving scenes, comprising three core components:
an unsupervised feature separator that isolates dynamic-static elements without manual labels, 
a tri-path deformable Mamba that models global context while preserving feature separation, 
and an adaptive gated fusion mechanism for robust, context-aware aggregation.
\item Our framework makes a pioneering contribution to LiDAR world models by introducing optional autonomous rollout and controllable generation capabilities which are previously unattainable by previous methods.
\item Extensive experiments on various benchmarks demonstrate that \mtd{} achieves state-of-the-art performance, validating the effectiveness of its design.
\end{itemize}

%% file: sec/2_related.tex
\section{Related Works}
\label{sec: related work}

\noindent\textbf{World Models in Autonomous Driving.}
Given historical observations, a world model aims to predict future states.
It is gaining growing attention in autonomous driving for its capability to generate high fidelity  data~\cite{zhao2025drivedreamer4d} and enhance driving safety~\cite{shi2025drivex,li2025end}.
In autonomous driving, three mainstream world models have emerged: vision-based~\cite{hu2023gaia, wang2024drivedreamer, wang2024driving}, occupancy-based~\cite{zheng2024occworld, shi2025come, gu2024dome}, and LiDAR-based~\cite{weng2021inverting,weng2022s2net,khurana2023point,zhang2023copilot4d,liang2025lidarcrafter}.
While former two types of world models are widely explored, the LiDAR counterpart remain less studied despite their capability to provide precise geometric comprehension and greater scalability.
Among these, several works~\cite{weng2021inverting,liang2025lidarcrafter,weng2022s2net} utilize sequence modeling techniques to predict future states.
4D-Occ~\cite{khurana2023point} employs voxel as an intermediate representation, and Copilot4D~\cite{zhang2023copilot4d} enhances model performance through compressed BEV feature interaction.
While existing methods overlook the properties of LiDAR data, we addresses this gap through meticulous designs, achieving not only more accurate prediction but also autonomous rollout and controllable generation.

\noindent\textbf{LiDAR Data Generation.}
Acquiring large-scale LiDAR data is time-consuming and costly~\cite{caesar2020nuScenes, bijelic2020seeing}, spurring growing research into LiDAR generation as a promising alternative.
Earlier methods~\cite{dosovitskiy2017carla, hahner2021fog, hahner2022lidar, kilic2025lidar} construct virtual scenes or physical models for generation, but they suffer from limited diversity and fidelity due to reliance on costly assets and hand-tuned parameters.
Recently, learning-based generative methods~\cite{nakashima2021learning,zyrianov2022learning,ran2024towards,nakashima2023lidar,wu2024text2lidar} directly model the distribution of LiDAR data, enabling large-scale synthesis and controllable sampling. 
While existing methods can produce high-fidelity single frame data, they fail to maintain temporal consistency across sequential scans. 
In contrast, our method can generate LiDAR sequences that are not only visually realistic but also temporally coherent.

\begin{figure*}[!t]
\centering
\includegraphics[width=0.98\textwidth]{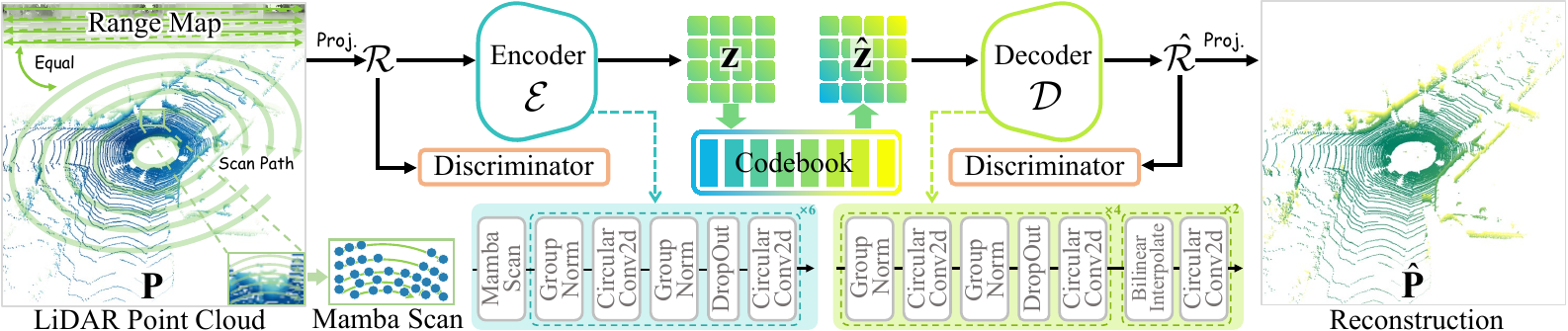}
\caption{
    Our \moduleone{} first projects the point cloud $\mathbf{P}$ into a dense range map $\mathcal{R}$. 
    A scan-based Mamba encoder then process the map into latent features $\mathbf{z}$, which are discretized into $\hat{\mathbf{z}}$ via a vector quantization codebook.
    After this, a decoder is used to perform the upsampling and reconstruct $\hat{\mathbf{z}}$ into output $\hat{\mathcal{R}}$, which is finally converted back into a point cloud $\hat{\mathbf{P}}$. 
    To enhance the reconstruction quality, particularly for range maps dominated by low-frequency signals, a discriminator is incorporated into our training process.
}
\label{fig: tokenizer}
\end{figure*}

\noindent\textbf{LiDAR Representation and Learning.}
Selecting an appropriate representation is crucial for LiDAR data generation.
While raw point clouds provide an intuitive data format, the irregular nature forces methods like PointNet~\cite{qi2017pointnet,qi2017pointnet++} to devote approximately 90\% of runtime to data organization rather than meaningful computation~\cite{shi2022pillarnet,liu2019point}, rendering it impractical for large-scale LiDAR generation.
As an alternative, voxel representation structures point clouds into volumetric grids, enabling feature extraction through CNN~\cite{maturana2015voxnet,qu2025robust,yan2018second,Zhu_2025_CVPR}.
However, this approach suffers from quantization error that compromise generation quality.
In contrast, range map representation offers a balanced solution.
Derived directly from LiDAR's polar output, this 2.5D format preserves geometric precision while maintain structural regularity.
Although compatible with CNN, we argue that  the range map's inherent structure aligns more naturally with Mamba's sequential scanning, enabling more effective feature learning for LiDAR generation.

%% file: sec/3_method.tex
\section{Methodology}
\label{sec:method}
This section first describes the problem formulation for our LiDAR world model in \cref{sec: formulation}. 
We then detail our dedicated tokenizer for LiDAR data compression  in \cref{sec: tokenizer}, and finally present our LiDAR world model with dynamic-static disentanglement in \cref{sec: wm}.

\subsection{Problem Formulation}
\label{sec: formulation}
World models, which predict future sensor data based on historical observations,  are becoming an increasingly prominent technique in fields like autonomous driving~\cite{kong20253d} and embodied intelligence~\cite{long2025survey}.
Formally, at a given timestamp $u$, a world model takes $\tau_p$ historical observations $\mathcal{P}^{p} = \{\mathbf{P}^{u}, \mathbf{P}^{u-1}, \cdots, \mathbf{P}^{u-\tau_p+1}\}$ and their corresponding ego status $\mathbf{a}^{p}$ as input, with the primary objective of predicting $\tau_f$ future frames $\mathcal{P}^{f} = \{\mathbf{P}^{u+1}, \mathbf{P}^{u+2}, \cdots, \mathbf{P}^{u+\tau_f}\}$.

We note two key variations in previous methods. 
First, future ego status $\mathbf{a}^f$ may be provided as ground-truth~\cite{weng2021inverting,khurana2023point,zhang2023copilot4d,gu2024dome}, predicted by a separate planner~\cite{weng2021inverting,shi2025come}, or generated by the world model itself~\cite{zheng2024occworld}. 
Second, some models can accept additional control signals (\eg, BEV layouts) to enable controllable generation, a feature of significant practical value for scenario-based testing. 
Inspired by recent latent diffusion-based approaches~\cite{shi2025come,gu2024dome} that successfully integrate these versatile functions, we similarly build our \mtd{} framework following a latent diffusion paradigm.

\noindent\textbf{Overall Pipeline.} 
The foundation of our framework is the \moduleone{} described in \cref{sec: tokenizer}, which serves two key purposes: it maps unordered point clouds into organized feature sequences and significantly reduces the computational cost of the subsequent diffusion process. 
This module consists of an encoder $\mathcal{E}$ that compresses point clouds into compact latent features and a decoder $\mathcal{D}$ that reconstructs LiDAR data from these features.

We first tokenize the historical point clouds $\mathcal{P}^{p}$ using $\mathcal{E}$ to obtain latent features $\mathcal{Z}^p \in \mathbb{R}^{\tau_p \times h \times w \times C}$. 
Following the approach of COME~\cite{shi2025come}, control signals, including historical and future ego status along with optional BEV layouts if provided, are encoded into control features.
All features are processed by our Mamba-based world model detailed in \cref{sec: wm}, producing an output $\mathcal{\hat Z}^f \in \mathbb{R}^{\tau_f \times h \times w \times C}$, which is then decoded by $\mathcal{D}$ into the final predicted point clouds.
Next, we details our LiDAR scene tokenizer and Mamba-based world model. 

\begin{figure*}[!t]
\centering
\includegraphics[width=0.98\textwidth]{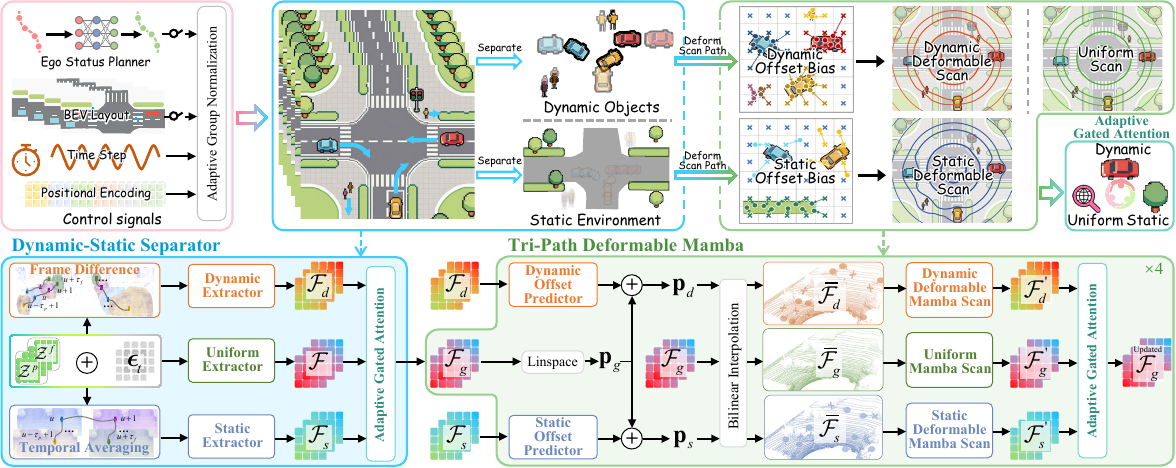}
\caption{
    The process of feature disentanglement and aggregation in \mtd{}.
    First, \moduletwo{} separates the latent features in an unsupervised manner into three components: dynamic features $\mathcal{F}_d$, static features $\mathcal{F}_s$, and generic features $\mathcal{F}$.
    The \modulethree{} then adaptively generates specialized scan paths $\{\mathbf{p}_{g}, \mathbf{p}_d, \mathbf{p}_s\}$ for each branches, which guide the sampling of intermediate features $\{\mathcal{\bar F}_{g}, \mathcal{\bar F}_d, \mathcal{\bar F}_s\}$ via bilinear interpolation.
    These features are then passed by Mamba within their respective branch, obtaining $\{\mathcal{F}_{g}^{'}, \mathcal{F}_{d}^{'}, \mathcal{F}_{s}^{'}\}$.
    Finally, an adaptive gated attention fuses these features to produce the refined feature $\mathcal{F}_{g}$ for future frame generation.
}
\vspace{-0.12cm}
\label{fig: flowchart}
\end{figure*}

\subsection{LiDAR Scene Tokenizer}
\label{sec: tokenizer}

The unordered and sparse nature of point clouds presents a significant challenge for effective feature extraction. 
We identify Mamba~\cite{gu2023mamba} as a highly suitable architecture for this task, as its core mechanism of feature aggregation through sequential scanning closely mirrors the physical process of LiDAR laser sweeps. 
Building on this observation, we propose the \moduleone{} for compact data representation, as illustrated in \cref{fig: tokenizer}.
Specifically, we first convert the input LiDAR point cloud $\mathbf{P}$ into a range map $\mathcal{R}\in \mathbb{R}^{H\times W}$ via spherical projection~\cite{milioto2019rangenet}, where $H$ and $W$ correspond to the number of vertical lasers and the horizontal resolution per scan, respectively. 
Each element in $\mathcal{R}$ stores the distance for a specific laser at a specific azimuth.
The $\mathcal{R}$ is then processed by a Mamba-based encoder $\mathcal{E}$, which consists of the initial Mamba scan operation followed by six blocks of common layers, as shown in \cref{fig: tokenizer}. 
As our implementation uses the standard Mamba operation, we omit its internal details and direct the readers to the paper~\cite{gu2023mamba} or our Appendix for more information.
With the spatial down-sampling occurs in the last block of the block, we can get the latent code $\mathbf{z}\in \mathbb{R}^{h \times w \times C}$.

During reconstruction, the latent representation $\mathbf{z}$ is first quantized via a codebook to obtain $\hat{\mathbf{z}}$. 
This quantized latent is then passed through a decoder $\mathcal{D}$, shown in \cref{fig: tokenizer}, to reconstruct the range map $\hat{\mathcal{R}} \in \mathbb{R}^{H\times W}$.
A vector-quantized reconstruction loss is utilized for model training: 
\begin{equation}
\mathcal{L}_{\mathtt{VQ}} = \mathbb{E}_{\mathcal{R}}||\mathcal{R}-\hat{\mathcal{R}}|| + \beta \cdot ||\mathtt{sg}(\mathbf{z})-\hat{\mathbf{z}}|| + ||\mathbf{z}-\mathtt{sg}(\hat{\mathbf{z}})||,
\end{equation}
where $\beta$ is a balancing hyperparameter and $\mathtt{sg}(\cdot)$ denotes the stop-gradient operation during training.

In $\mathcal{L}_{\mathtt{VQ}}$, the mean-square error tends to produce overly smooth predictions for nearby pixels.
However, the accurate reconstruction of such high-frequency information is critical for range maps, as it corresponds to essential geometric features like object edges and scene structures~\cite{wu2024text2lidar}.
To address this spectral bias and enhance reconstruction fidelity, we introduce a discriminator $\mathcal{S}$~\cite{ran2024towards} with an extra adversarial training objective, which is illustrated in \cref{fig: tokenizer}:
\begin{equation}
\begin{split}
\mathcal{L}_{\mathtt{ADV}} = \mathbb{E}_{\mathcal{R}}[\mathtt{log}(\mathcal{S}(\mathcal{R})) + \mathtt{log}(1 - \mathcal{S}(\hat{\mathcal{R}}))],
\end{split}
\end{equation}
and the whole loss for our \moduleone{} is:
\begin{equation}
\mathcal{L}_{\mathtt{LST}} = \mathcal{L}_{\mathtt{VQ}} + \mathcal{L}_{\mathtt{ADV}}.
\end{equation}

\subsection{Deformable Mamba for LiDAR World Model}
\label{sec: wm}
This section first details our Mamba-based feature extractor, which comprises the \moduletwo{} and \modulethree{} as shown in \cref{fig: flowchart}. 
We then outline the process for generating future frames from enhanced features and conclude with an overview of controllable generation.

\noindent\textbf{Dynamic-Static Separator.}
Compared to camera videos and occupancy data, LiDAR data provides relatively weaker semantic cues for scene understanding, which hinders world models from efficiently comprehending environmental evolution. 
This limitation motivates us to explicitly separate dynamic objects from the static environment to enhance the model's representational capacity. 
To avoid labor-intensive semantic labeling and improve scalability, we propose to accomplish this disentanglement in an unsupervised manner.

Leveraging the observation that dynamic objects exhibit variation across frames while static environments remain largely consistent, we derive two complementary cues: frame difference and temporal averaging. 
For brevity, the denoising timestep $t$ is omitted in the feature expressions below.
Given a latent feature $\mathcal{Z} \in \mathbb{R}^{\tau \times h \times w \times C}$, we compute the dynamic pattern $\mathcal{Z}_d$ via frame-wise differences:
\begingroup
\small
\begin{equation}
    \mathcal{Z}_d[i] = 
    \begin{cases}
    \mathbf{0}, & \text{if } i = 1,\\
    \mathcal{Z}[:,i] - \mathcal{Z}[:,i-1],  & \text{if } i = 2, 3, \cdots, \tau.
    \end{cases}
\end{equation}
\endgroup

For the static pattern, we apply a temporal average within a sliding window of size $n$ to obtain $\mathcal{Z}_s$:
\begingroup
\small
\begin{equation}
\begin{split}
\mathcal{Z}_s[i] = \frac{1}{\mathtt{end}-\mathtt{start}}\sum\nolimits^{\mathtt{end}-1}_{i=\mathtt{start}}\mathcal{Z}[i], \ i=1, 2, \cdots, \tau,
\end{split}
\end{equation}
\endgroup
where $\mathtt{start} = \mathtt{max}(0, i-\lfloor n/2\rfloor),
\mathtt{end} = \mathtt{min}(\tau, i+\lfloor n/2\rfloor +1)$ define the window bounds. 

Subsequently, we employ three parallel extractors with identical 3D convolutional structures to transform ${\mathcal{Z}, \mathcal{Z}_d, \mathcal{Z}_s}$ into features ${\mathcal{F}, \mathcal{F}_d, \mathcal{F}_s}$. 
To effectively fuse these features, we introduce an adaptive gated attention inspired by~\cite{qiu2025gated, wu2023co}, which performs data-dependent fusion:
\begin{equation}
\begin{split}
\mathcal{F}^{'} &= (1-\mathtt{G}(\mathtt{C}[\mathcal{F}_{d},\mathcal{F}_{s}]))\cdot \mathcal{F}_{d} + \mathtt{G}(\mathtt{C}[\mathcal{F}_{d},\mathcal{F}_{s}])\cdot \mathcal{F}_{s}, \\
\mathcal{F}_{g} &= (1-\mathtt{G}(\mathcal{F}'))\cdot \mathcal{F} + \mathtt{G}(\mathcal{F}^{'})\cdot \mathcal{F}^{'},
\end{split}
\end{equation}
where $\mathtt{G}(\cdot)$ is a gating function composed of convolutional layers and activations, and $\mathtt{C}[\cdot,\cdot]$ denotes channel-wise concatenation.
Ultimately, we have $\{\mathcal{F}_d, \mathcal{F}_s\}$ for dynamic and static cues, along with a fused generic feature $\mathcal{F}_{g}$.

\begin{figure}[!t]
\includegraphics[width=0.478\textwidth]{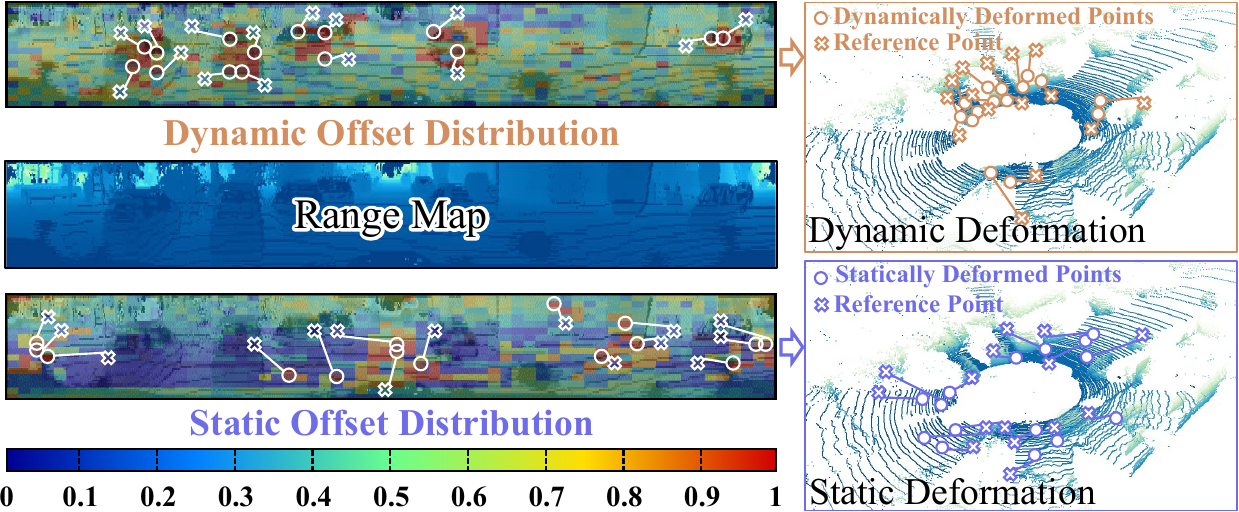}
\caption{Visualization of scan path deformation on range maps and LiDAR point clouds. Warmer colors in the heatmap indicate a higher concentration of path points shifted to those regions. We visualize the displacement trends of the top-15 most frequently deformed path points.}
\vspace{-0.1cm}
\label{fig: dynamic_static}
\end{figure}

\noindent\textbf{Tri-path Deformable Mamba.}
After obtaining dynamic and static cues, we propose a tri-path architecture to leverage them for disentangled region modeling while maintaining global coherence.
We begin with a brief overview of deformable Mamba scanning and interpolated feature generation for later clarity.
Given an input generic feature $\mathcal{F}_{g} \in \mathbb{R}^{\tau\times h\times w \times C}$ together with customized scan paths $\mathbf{p}_{\gamma} \in \mathbb{R}^{(\tau\times h\times w) \times 3}$, where $\gamma\in \{g, d, s\}$ denotes the generic, dynamic, and static paths, the deformable Mamba mechanism~\cite{liu2025defmamba} aims to aggregate features along this path and refines them into the output feature $\mathcal{F}^{'}_{\gamma}$:
\begin{equation}
\mathcal{\bar F}_{\gamma} = \mathtt{BI}(\mathcal{F}_{g}, \mathbf{p}_{\gamma}), \quad \mathcal{F}^{'}_{\gamma} = \mathtt{DM}(\mathcal{\bar F}_{\gamma}),
\quad \gamma \in \{ g, d, s \}
\label{eq:defmamba}
\end{equation}
where $\mathtt{BI}(\mathcal{F}_{g}, \mathbf{p}_{\gamma})$ denotes bilinear interpolation for sampling $\mathcal{F}_{g}$ along path $\mathbf{p}_{\gamma}$, and $\mathtt{DM}(\cdot)$ denotes performing Mamba operation~\cite{gu2023mamba} at the deformed feature.

Specifically, as illustrated in \cref{fig: flowchart}, each branch in our \modulethree{} performs the two-step procedure defined in \cref{eq:defmamba}. 
The first step involves feature aggregation via a branch-specific reference path $\mathbf{p}_{\gamma}$. 
For the generic branch, we directly use the standard path $\mathbf{p}_{g} \in \mathbb{R}^{(\tau\times h\times w) \times 3}$ obtained from the 3D coordinate space $\mathbb{C}$:
\begin{equation}
\mathbf{p}_{g} = \mathtt{Flatten}(\mathtt{Meshgrid}(\mathtt{Linspace}(\mathbb{C}))).
\end{equation}
For the dynamic and static branches, however, the model need to learn to focus on the corresponding elements. 
A key challenge is customizing these paths effectively. 
We address this by leveraging the pre-computed features ${\mathcal{F}_d, \mathcal{F}_s}$ as indicators for the locations of dynamic and static elements. 
These features are processed through fully connected layers to generate specialized paths:
\begin{equation}
    \begin{split}
        \mathbf{p}_{d} = \mathbf{p}_{g} + \mathtt{Tanh}(\mathtt{Linear}(\mathtt{ReLu}(\mathtt{Linear}(\mathcal{F}_{d})))), \\
        \mathbf{p}_{s} = \mathbf{p}_{g} + \mathtt{Tanh}(\mathtt{Linear}(\mathtt{ReLu}(\mathtt{Linear}(\mathcal{F}_{s})))).
    \end{split}
\end{equation}
Guided by these learned paths, the scanner can direct its attention toward either dynamic objects or the static background, effectively separating the respective information, which is verified by our visualizations in \cref{fig: dynamic_static}.
Using these three distinct paths, we obtain interpolated features $\{\mathcal{\bar F}_{g}, \mathcal{\bar F}_d, \mathcal{\bar F}_s\}$, with each deforming to different regions of the driving scene.
Subsequently, the three features are processed by Mamba within their respective branches, obtaining $\{\mathcal{F}_{g}^{'}, \mathcal{F}_{d}^{'}, \mathcal{F}_{s}^{’}\}$, with each capturing different aspects of the driving scene.
The outputs from these three branches are then fused via our adaptive gated attention mechanism to produce an updated generic feature $\mathcal{F}_{g}$, completing one processing block of \modulethree{}.
In our implementation, we stack four such blocks, which iteratively refine $\mathcal{F}_{g}$ for the final generation task.

\noindent\textbf{Latent Diffusion Paradigm.} 
We employ the diffusion paradigm~\cite{rombach2022high, peebles2023scalable} for generation.
During training, the target future point clouds are encoded into latent features $\mathcal{Z}^f\in \mathbb{R}^{\tau_f\times h \times w \times C}$.
Over $T$ diffusion steps, noise is progressively added to $\mathcal{Z}^{f}$ at each step $t$ to form a noisy latent $\mathcal{Z}^{f}_t = \sqrt{\alpha_{t}}\mathcal{Z}^{f} +  \sqrt{1 - \alpha_{t}}\bm{\epsilon}_{t}$, where $\alpha_{t}$ is a noise schedule weight that decreases with $t$, and $\bm{\epsilon}_{t}$ is random noise. 
In the denoising stage, the corresponding refined $\mathcal{F}_{g}$ at each timestamp $t$ is processed by a convolutional layer to get the predicted noise $\hat{\bm{\epsilon}}_{t}$.
The training objective of our latent diffusion process is to minimize the error between predicted and actual noise, formulated as:
\begin{equation}
\mathcal{L}_{\mathtt{LD}}(\bm{\epsilon}_{t}, \hat{\bm{\epsilon}}_{t})
= \mathbb{E}_{\bm{\epsilon} \sim \mathcal{N}(\mathbf{0},\mathbf{1})} [||\bm{\epsilon}_{t} - \mathtt{WM}(\mathcal{Z}^f_{t}, \mathcal{Z}^p_{t}, \mathbf{t},\mathbf{c};\bm{\theta}_{\mathtt{WM}})||^{2}],
\label{eq: wmloss}
\end{equation}
where $\bm{\theta}_{\mathtt{WM}}$ denotes the learnable parameters of the world model, and $\textbf{t}$ is the timestamp embedding. 
$\textbf{c}$ is the optional control signals introduced below. 

After training, the model synthesizes future latents through an iterative denoising process.
This begins with an initial noise latent $\mathcal{\hat Z}^f_0 \sim \mathcal{N}(\mathbf{0}, \mathbf{I})$ and progressively recover it over $T$ steps to produce the final prediction.

\noindent\textbf{Controllable Generation.}
In real-world applications, future ego status $\mathbf{a}^f$ are typically unknown in advance, which is a critical issue overlooked by previous methods.
To address this, we incorporate an optional planner to predict these status. Inspired by BEV-Planner~\cite{li2024ego}, our planner is jointly learned with world model to fully leverage perceptual and ego status information as:
\begin{equation}
\mathcal{L}_{\mathtt{Planner}}(\mathbf{a}^{f},  \hat{\mathbf{a}}^{f})
= ||\mathbf{a}^{f} - \mathtt{Planner}(\mathbf{a}^{p};\bm{\eta}|\bm{\theta}_{\mathtt{WM}})||^{2},
\label{eq: planloss}
\end{equation}
where $\mathbf{a}^{f}$ is the ground truth ego status, $\hat{\mathbf{a}}^{f}$ is the predicted ego status, and $\bm{\eta}$ is the parameters of the planner.

Following established practices~\cite{shi2025come,gu2024dome}, we encode timestamps and ego status, along with optional signals like BEV layouts, into feature embeddings. These are fused via adaptive group normalization~\cite{dhariwal2021diffusion} to form the conditioning vector $\mathbf{c}$ in \cref{eq: wmloss}. 
This design offers significant flexibility through a unified architecture that accommodates various control signals types and sources.
The complete training objective for our world model is:
\begin{equation}
\mathcal{L} = \mathcal{L}_{\mathtt{LD}} + \mathcal{L}_{\mathtt{Planner}}.
\end{equation}

%% file: sec/4_exp.tex
\begin{figure*}[!t]
\centering
\includegraphics[width=0.937\textwidth]{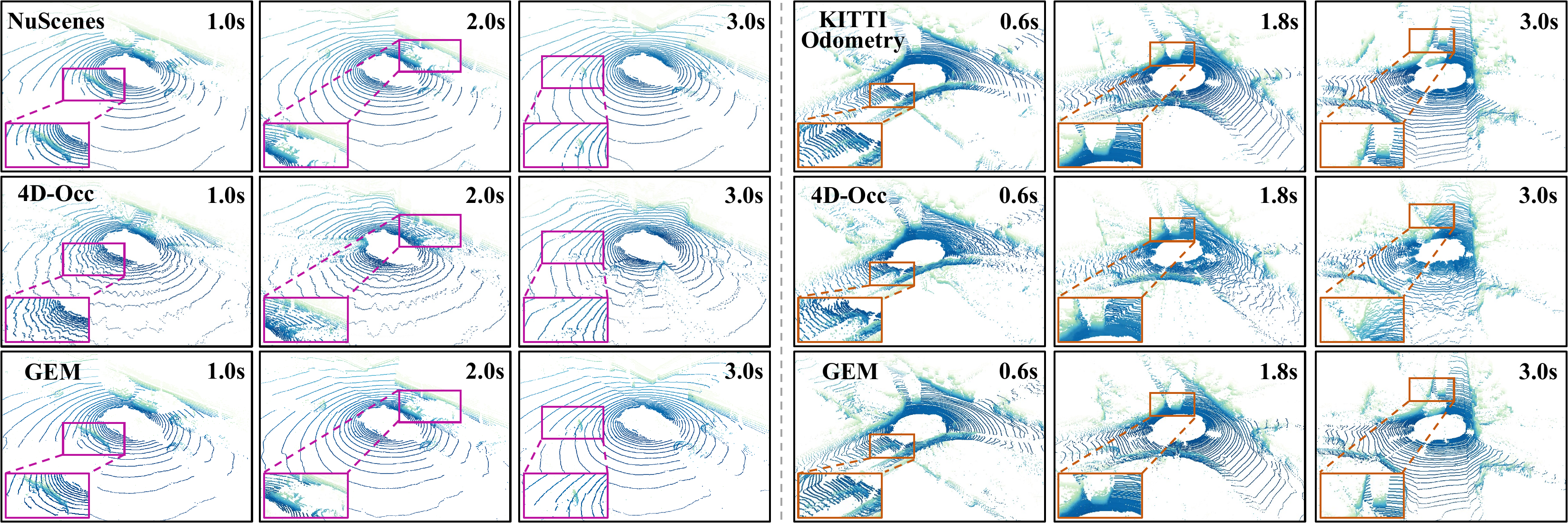}
\caption{Visual comparisons with competitive LiDAR world models on nuScenes~\cite{caesar2020nuScenes} and KITTI Odometry~\cite{geiger2012we} datasets.}
\label{fig: visual_nuscenes_kitti}
\end{figure*}
\section{Experiments}
\label{sec:exp}
\begin{table}[!t]
\Huge
\centering
\renewcommand\arraystretch{1.15}
\caption{Quantitative comparisons of world modeling accuracy  with other state-of-the-art methods on nuScenes~\cite{caesar2020nuScenes}.}
\label{tab: nuscenes}
\resizebox{0.478\textwidth}{!}{
\begin{tabular}{lccccccc}
\toprule[1.2pt]
Methods & Horizon & CD$_{inner}$ $\downarrow$ & L1$_{sr} \downarrow$ & AbsRel$_{sr} \downarrow$ & L1 $\downarrow$ & AbsRel $\downarrow$ & CD $\downarrow$ \\ \hline
\multirow{2}{*}{SPFNet} & 1s & 2.24 & - & - & 4.58 & 34.87 & 4.17 \\
& 3s & 2.50 & - & - & 5.11 & 32.74 & 4.14 \\ \arrayrulecolor{gray!13}\hline\arrayrulecolor{black}
\multirow{2}{*}{S2Net}   & 1s & 1.70 & - & - & 3.49 & 28.38 & 2.75 \\
& 3s & 2.06 & - & - & 4.78 & 30.15 & 3.47 \\ \arrayrulecolor{gray!13}\hline\arrayrulecolor{black}
\multirow{2}{*}{Ray tracing} & 1s & 0.54 & - & - & 1.50 & 14.73 & 0.90 \\
& 3s & 1.66 & - & - & 2.44 & 26.86 & 3.59 \\ \arrayrulecolor{gray!13}\hline\arrayrulecolor{black}
\multirow{2}{*}{4D-Occ}  & 1s & 1.41 & \underline{0.81} & \underline{0.61} & 1.40 & 10.37 & 2.81 \\
& 3s & 1.40 & \underline{0.75} & \underline{0.49} & 1.71 & 13.48 & 4.31 \\ \arrayrulecolor{gray!13}\hline\arrayrulecolor{black}
\multirow{2}{*}{Copilot4D} & 1s & \underline{0.36} & 0.92 & 0.85 & \underline{1.30} & \underline{8.58} & \underline{2.01} \\
& 3s & \underline{0.58} & 0.91 & 0.82 & \underline{1.51} & \textbf{10.38} & \underline{2.47} \\ \arrayrulecolor{gray!13}\hline\arrayrulecolor{black}
\rowcolor{green!7}
& 1s & \textbf{0.30} & \textbf{0.01} & \textbf{0.07} & \textbf{0.98} & \textbf{6.67} & \textbf{0.38} \\
\rowcolor{green!7}
\multirow{-2}{*}{\textbf{Ours}} & 3s & \textbf{0.51} & \textbf{0.08} & \textbf{0.19} & \textbf{1.43} & \underline{11.13} & \textbf{0.67} \\ \bottomrule[1.2pt]
\end{tabular}
}
\vspace{-0.17cm}
\end{table}
\begin{table}[!t]
\Huge
\centering
\renewcommand\arraystretch{1.15}
\caption{Quantitative comparisons of world modeling accuracy with other state-of-the-art methods on KITTI Odometry~\cite{geiger2012we}.}
\label{tab: kitti-o}
\resizebox{0.478\textwidth}{!}{
\begin{tabular}{lccccccc}
\toprule[1.2pt]
Methods & Horizon & CD$_{inner}$ $\downarrow$ & L1$_{sr} \downarrow$ & AbsRel$_{sr} \downarrow$ & L1 $\downarrow$ & AbsRel $\downarrow$ & CD $\downarrow$  \\ \midrule
\multirow{2}{*}{ST3DCNN} & 1s & 4.11 & - & - & 3.13 & 26.94 & 4.51 \\
& 3s & 4.19 & - & - & 3.25 & 28.58 & 4.83 \\ \arrayrulecolor{gray!13}\hline\arrayrulecolor{black}
\multirow{2}{*}{Ray tracing} & 1s & 0.62 & - & - & 1.50 & 16.15 & 0.76 \\
& 3s & 4.01 & - & - & 2.82 & 29.67 & 5.92 \\ \arrayrulecolor{gray!13}\hline\arrayrulecolor{black}
\multirow{2}{*}{4D-Occ}  & 1s & 0.51 & \underline{0.82} & \underline{0.72} & 1.12 & 9.09 & 0.61 \\
& 3s & 0.96 & \underline{0.78} & \underline{0.54} & 1.45 & 12.23 & 1.50 \\ \arrayrulecolor{gray!13}\hline\arrayrulecolor{black}
\multirow{2}{*}{Copilot4D} & 1s & \underline{0.18} & 0.88 & 0.85 & \underline{0.95} & \underline{8.59} & \underline{0.21} \\
& 3s & \underline{0.45} & 0.87 & 0.81 & \underline{1.27} & \underline{11.50} & \underline{0.67} \\ \arrayrulecolor{gray!13}\hline\arrayrulecolor{black}
\rowcolor{green!7}
& 1s & \textbf{0.13} & \textbf{0.15} & \textbf{0.13} & \textbf{0.81} & \textbf{7.08} & \textbf{0.17} \\
\rowcolor{green!7}
\multirow{-2}{*}{\textbf{Ours}} & 3s & \textbf{0.24} & \textbf{0.10} & \textbf{0.09} & \textbf{1.11} & \textbf{10.44} & \textbf{0.28} \\ \bottomrule[1.2pt]
\end{tabular}
}
\vspace{-0.17cm}
\end{table}

In this part, we detail the experimental settings in \cref{sec:datasets}, present quantitative and qualitative comparison in \cref{sec:exp_world,sec:exp_single}, provide extensive ablation studies in \cref{sec:abltaion}, and show comprehending and modeling capacity in \cref{sec:world_comprehend}.

\subsection{Datasets and Evaluation Metrics}
\label{sec:datasets}
\noindent\textbf{Implementation Details.}
\mtd{} is trained in two stages: (1) firstly, \moduleone{} is trained for 80 epochs with Adam~\cite{kingma2014adam} and a learning rate of $4e-4$;
(2) secondly, LiDAR model is trained for 1.2M steps using AdamW~\cite{loshchilov2017decoupled} with a learning rate of $2e-4$.  
Unless otherwise specified, all models are trained on 8 H20 GPUs.
We evaluate the performance of \mtd{} from two different perspectives: future prediction accuracy and generated distribution realism.

\noindent\textbf{For LiDAR World Modeling Accuracy.} 
We conduct experiments on two widely adopted autonomous driving datasets: nuScenes~\cite{caesar2020nuScenes} and KITTI Odometry~\cite{geiger2012we}. 
Our method is compared against six state-of-the-art approaches: SPFNet~\cite{weng2021inverting}, S2Net~\cite{weng2022s2net}, ST3DCNN~\cite{mersch2022self}, Ray Tracing~\cite{khurana2023point}, 4D-Occ~\cite{khurana2023point}, and Copilot4D~\cite{zhang2023copilot4d}. 
For fair comparisons on the nuScenes benchmark, we evaluate two forecasting horizons: 1-second and 3-second, using 2 and 6 historical frames, respectively.
On KITTI Odometry, we follow the standard protocol of using 5 historical frames to predict 5 future frames. 
For comprehensive evaluations, we adopt four established metrics: Chamfer Distance (CD) for overall geometric consistency, Inner-CD (CD$_{inner}$) for near-range accuracy, L1 Depth Error (L1) for absolute depth error, and Absolute Relative Error (AbsRel) for relative depth precision. 
Considering the importance of prediction stability, we introduce two stability metrics.
Specifically, we define AbsRel$_{sr}$ and 
L1$_{sr}$ as the absolute difference between $1$ and the ratio of the median to the mean of their respective error distributions. 
A value closer to $0$ indicates more stable and reliable predictions.

\noindent\textbf{For Generated Distribution Realism.} 
We evaluate the realism of generated data against 4D-Occ on the nuScenes and KITTI Odometry datasets. 
For a broader assessment, we also compare with single-frame methods on the larger-scale KITTI-360 dataset~\cite{liao2022kitti}, including UltraLiDAR~\cite{xiong2023learning}, LiDM~\cite{ran2024towards}, Text2LiDAR~\cite{wu2024text2lidar}, and WeatherGen~\cite{wu2025weathergen}. 
Following LiDM, we use 2,000 randomly generated samples and five metrics.
Three F\'{r}echet distances including FRID, FSVD and FPVD are utilized to quantify the feature distribution discrepancy by leveraging features from RangeNet++~\cite{milioto2019rangenet}, MinkowskiNet~\cite{choy20194d}, and SPVCNN~\cite{tang2020searching} at the range image, sparse volume, and point-based volume levels, respectively.
Additionally, Jensen–Shannon Divergence (JSD) and Minimum Matching Distance (MMD) are employed to assess the geometric distribution realism.

\subsection{LiDAR World Modeling Analysis}
\label{sec:exp_world}
\cref{tab: nuscenes,tab: kitti-o} show quantitative comparisons with other state-of-the-art methods~\cite{weng2021inverting, weng2022s2net, mersch2022self, khurana2023point, zhang2023copilot4d} on nuScenes and KITTI Odometry datasets, and the results demonstrate that our method outperforms all others across two datasets, achieving state-of-the-art performance. 
For example, on the nuScenes for 1s prediction, our method achieves the best results across all metrics, particularly reducing the CD by 81.1$\%$ compared to the second-best method. 
For 3s prediction, our method slightly underperforms Copilot4D on the AbsRel metric but leads in all other metrics. 
On KITTI Odometry, our method comprehensively outperforms competing approaches for both 1s and 3s predictions.
Additionally, unlike other methods that suffer from performance instability over time, our method benefits from disentangled dynamic-static modeling to achieve excellent stable performance, which is supported by lower L1$_{sr}$ and AbsRel$_{sr}$ values.
For qualitative evaluation, \cref{fig: visual_nuscenes_kitti} presents a visual comparison with the open-sourced 4D-Occ. It can be observed that our method outperforms others in both detail fidelity and temporal consistency.
Furthermore, our method demonstrates speed advantage over 4D-Occ, underscoring its greater practical value. 
When benchmarked on a 4090 GPU, \mtd{} achieves 9.23 FPS on nuScenes and 4.67 FPS on KITTI Odometry, outperforming 4D-Occ reaching only 7.41 FPS and 3.21 FPS on the respective datasets.

\subsection{Generated Distribution Analysis}
\label{sec:exp_single}
\begin{table}[!t]
\Huge
\centering
\renewcommand\arraystretch{1.15}
\caption{Quantitative comparisons of generated distribution realism  with other state-of-the-art methods on three datasets.}
\label{tab: three}
\resizebox{0.478\textwidth}{!}{
\begin{tabular}{clccccc}
\toprule[0.8pt]
Datasets & Methods & FRID $\downarrow$ & FSVD $\downarrow$ & FPVD $\downarrow$ & JSD $\downarrow$ & MMD $\times10^{-4}\downarrow$ \\ \hline
\multirow{2}{*}{nuScenes}
 & 4D-Occ & - & 30.4 & 23.2 & 0.308  & 3.52 \\
 & \cellcolor{green!7}\textbf{Ours} & \cellcolor{green!7}- & \cellcolor{green!7}\textbf{16.1} & \cellcolor{green!7}\textbf{13.0}  & \cellcolor{green!7}\textbf{0.095}  & \cellcolor{green!7}\textbf{2.45} \\
\arrayrulecolor{gray!13}\hline\arrayrulecolor{black}
\multirow{2}{*}{\makecell{KITTI \\ Odometry}}
 & 4D-Occ & 252.9 & 36.3 & 35.11 & \textbf{0.201} & \textbf{2.07} \\
 & \cellcolor{green!7}\textbf{Ours} & \cellcolor{green!7}\textbf{88.0}  & \cellcolor{green!7}\textbf{27.5} & \cellcolor{green!7}\textbf{27.3} & \cellcolor{green!7}\textbf{0.201}  & \cellcolor{green!7}2.18 \\
\arrayrulecolor{gray!13}\hline\arrayrulecolor{black}
\multirow{7}{*}{KITTI-360}
 & UltraLiDAR & 370.0 & 72.1 & 66.6 & 0.747 & 17.12 \\
 & LiDARGen & 129.1 & 39.2 & 33.4 & 0.188 & 2.88 \\
 & LiDM & 125.1 & 38.8 & 29.0 & 0.211 & 3.84 \\
 & Text2LiDAR & \textbf{72.3} & 47.8 & 40.6 & 0.132 & 3.00 \\
 & WeatherGen & 91.2 & 26.9 & \underline{23.1} & \underline{0.131} & \underline{2.59} \\
 & 4D-Occ & 164.2 & \underline{23.6} & 23.4 & 0.179 & 3.07  \\
 & \cellcolor{green!7}\textbf{Ours} &  \cellcolor{green!7}\underline{94.0} & \cellcolor{green!7}\textbf{23.3} & \cellcolor{green!7}\textbf{18.7} & \cellcolor{green!7}\textbf{0.125} & \cellcolor{green!7}\textbf{2.71} \\
\bottomrule[0.8pt]
\end{tabular}
}
\end{table}

Beyond paired ground-truth-based accuracy evaluation for predicted future frames, 
we also measure the distribution discrepancy between generated outputs and the real datasets.
\cref{tab: three} presents a comparison between our method and 4D-Occ on nuScenes and KITTI Odometry. 
It can be observed that our method outperforms 4D-Occ across nearly all metrics, suggesting that its outputs more closely resemble the real datasets and exhibit superior realism.
For broader validation, we also compare our approach with single-frame generation methods. 
As shown, \mtd{} achieves optimal performance on four out of the five metrics across both feature and geometric distributions, validating the high fidelity of each generated frame.
These results further demonstrate that \mtd{} not only achieve remarkable accuracy in world modeling but also produce more realistic data, which explore the possibility of generating large-scale, high-fidelity dataset in a cost-effective and scalable manner.

\subsection{Ablation Study}
\label{sec:abltaion}
\begin{table}[]
\Large
\centering
\renewcommand\arraystretch{1.15}
\caption{Ablation studies of architecture designs on nuScenes~\cite{caesar2020nuScenes} dataset under 3s setting.}
\label{tab: architecture}
\resizebox{0.478\textwidth}{!}{
\begin{tabular}{ccccccc}
\toprule[0.8pt]
Frameworks & CD$_{inner}$ $\downarrow$ & L1$_{sr} \downarrow$ & AbsRel$_{sr} \downarrow$ & L1 $\downarrow$ & AbsRel $\downarrow$ & CD $\downarrow$ \\ \hline
Unet & 0.81 & \textbf{0.07} & \underline{0.20} & 1.85 & 16.22 & 1.07 \\
\arrayrulecolor{gray!13}\hline\arrayrulecolor{black}
DiT & 0.66 & 0.11 & 0.23 & 1.69 & \textbf{10.67} & 0.90 \\
\arrayrulecolor{gray!13}\hline\arrayrulecolor{black}
Vision Mamba & 0.67 & 0.09 & \textbf{0.19} & 1.64 & 11.32 & 0.89 \\ 
\arrayrulecolor{gray!13}\hline\arrayrulecolor{black}
Triple Mamba & \underline{0.55} & 0.09 & 0.23 & \underline{1.49} & 12.54 & \underline{0.72} \\ 
\arrayrulecolor{gray!13}\hline\arrayrulecolor{black}
\rowcolor{green!7}
\textbf{Ours} & \textbf{0.51} & \underline{0.08} & \textbf{0.19} & \textbf{1.43} & \underline{11.13} & \textbf{0.67} \\
\bottomrule[0.8pt]
\end{tabular}
}
\end{table}
\begin{table}[!t]
\Huge
\centering
\caption{Ablation studies on the nuScenes~\cite{caesar2020nuScenes} dataset under 3s setting to investigate contribution of each key component of \modulethree{}: the Dynamic Extractor (DE), Static Extractor (SE), Dynamic Deformable Mamba (DDM), Static Deformable Mamba (SDM), and Adaptive Gated Attention (AGA).}
\renewcommand\arraystretch{1.15}
\label{tab: ablation}
\resizebox{0.478\textwidth}{!}{
\begin{tabular}{ccccccccccc}
\toprule[0.8pt]
\multicolumn{5}{c}{Key Designs} 
& \multirow{2}{*}{CD$_{inner}$ $\downarrow$}
& \multirow{2}{*}{L1$_{sr} \downarrow$}
& \multirow{2}{*}{AbsRel$_{sr} \downarrow$}
& \multirow{2}{*}{L1 $\downarrow$}
& \multirow{2}{*}{AbsRel $\downarrow$}
& \multirow{2}{*}{CD $\downarrow$} \\
DE & SE & SDM & DDM & AGA & & & & & & \\ \hline
& $\checkmark$ & $\checkmark$ &  & $\checkmark$ & 0.66 & 0.12 & 0.23 & 1.56 & 11.99 & 1.01 \\
$\checkmark$ & & & $\checkmark$ & $\checkmark$ & 0.53 & 0.10 & 0.19 & 1.44 & 11.35 & 0.78 \\ 
$\checkmark$ & $\checkmark$ & &  & $\checkmark$ & 0.76 & 0.08 & 0.22 & 1.68 & 12.82 & 1.17 \\ 
$\checkmark$ & $\checkmark$ & $\checkmark$ & $\checkmark$ &  & 0.57 & 0.09 & 0.19 & 1.47 & 11.62 & 0.88 \\ 
\rowcolor{green!7}
$\checkmark$ & $\checkmark$ & $\checkmark$ & $\checkmark$ & $\checkmark$ & \textbf{0.51} & \textbf{0.08} & \textbf{0.19} & \textbf{1.43} & \textbf{11.13} & \textbf{0.67} \\ 
\bottomrule[0.8pt]
\end{tabular}
}
\end{table}

\begin{table}[!t]
\large
\centering
\renewcommand\arraystretch{1.15}
\caption{A quantitative comparison of reconstruction quality with other architectures on KITTI-360~\cite{liao2022kitti}. Conv. , Trans. , and Dis. stand for convolution, transformer, and discriminator.}
\label{tab: recon}
\resizebox{0.478\textwidth}{!}{
\begin{tabular}{lcccccc}
\toprule[0.8pt]
\multirow{2}{*}{Methods} & \multirow{2}{*}{Frameworks} & \multicolumn{3}{c}{Perceptual} & \multicolumn{2}{c}{Statistical} \\  \cmidrule(r){3-5}\cmidrule(r){6-7}
& & FRID $\downarrow$ & FSVD $\downarrow$ & FPVD $\downarrow$ & CD $\downarrow$ & EMD $\downarrow$  \\ \hline
LiDM & Conv.+ Dis. & 0.40 & 11.2 & 12.2 & 0.094 & 0.199 \\
\arrayrulecolor{gray!13}\hline\arrayrulecolor{black}
\rowcolor{green!7}
\multirow{3}{*}{} & Trans.+Dis. & 0.21 & 7.9 & 9.7 & 0.021 & 0.080 \\
\rowcolor{green!7}
\textbf{Ours} & Mamba & \underline{0.20} & \underline{6.9} & \underline{8.3} & \underline{0.019} & \underline{0.075} \\
\rowcolor{green!7}
& Mamba+Dis.& \textbf{0.19} & \textbf{6.7} & \textbf{7.9} & \textbf{0.018} & \textbf{0.070} \\
\bottomrule[0.8pt]
\end{tabular}
}
\end{table}
\noindent\textbf{Effects of Different Model Architecture.}
To validate our architecture's effectiveness, we conduct ablation studies by replacing the \modulethree{} with UNet~\cite{nakashima2023lidar}, DiT~\cite{peebles2023scalable}, and Vision Mamba~\cite{zhu2024vision} and a triple Mamba variant of \mtd{} that uses standard Mamba in place of deformable Mamba. 
As shown in \cref{tab: architecture}, our design achieves top-tier performance across all metrics, ranking either first or second. Notably, comparison with the triple Mamba variant confirms that our performance gain stems from explicit dynamic-static modeling rather than increased parameters.

\noindent\textbf{Effects of Key Designs in Disentangled Modeling.}
We examine the impacts of different key design of disentangled modeling.
As shown in \cref{tab: ablation}, removing the dynamic extractor (DE) and dynamic deformable mamba (DDM) substantially degrades temporal consistency, with L1 and CD metrics deteriorating by 9.1$\%$ and 50.7$\%$ respectively.
Similarly, the removal of static extractor (SE) and static deformable mamba (SDM) also impairs performance.
More notably, the performance drop drastically when dynamic-static features are not used to guide the deformable Mamba, with L1 and CD decreasing by 17.5$\%$ and 74.6$\%$.
This demonstrates that explicit modeling is required to fully utilize these discriminative features. 
Furthermore, without adaptive gated attention (AGA) to fuse features across branches, AbsRel and CD decrease by 4.4$\%$ and 23.9$\%$. 
Together, these findings validate the importance of integrating all components for effective disentangled modeling.

\noindent\textbf{Reconstruction Performance of Different Tokenizer Architecture.}
\cref{tab: recon} compares the reconstruction performance against LiDM and other alternative on the KITTI-360 dataset from both perceptual and statistical perspectives.
Notably, the Mamba-only variant outperforms convolution- and transformer-based methods, which even employ discriminators for enhanced reconstruction. 
When similarly augmented with a discriminator, our approach achieves further performance gains.
Benefiting from the inherent alignment between range map structure and Mamba scanning, our architecture establishes a solid foundation for subsequent world modeling. 
Additional qualitative visualizations for comparison can be found in the Appendix.

\subsection{World Comprehending and Modeling}
\label{sec:world_comprehend}
\begin{figure}[!t]
\includegraphics[width=0.478\textwidth]{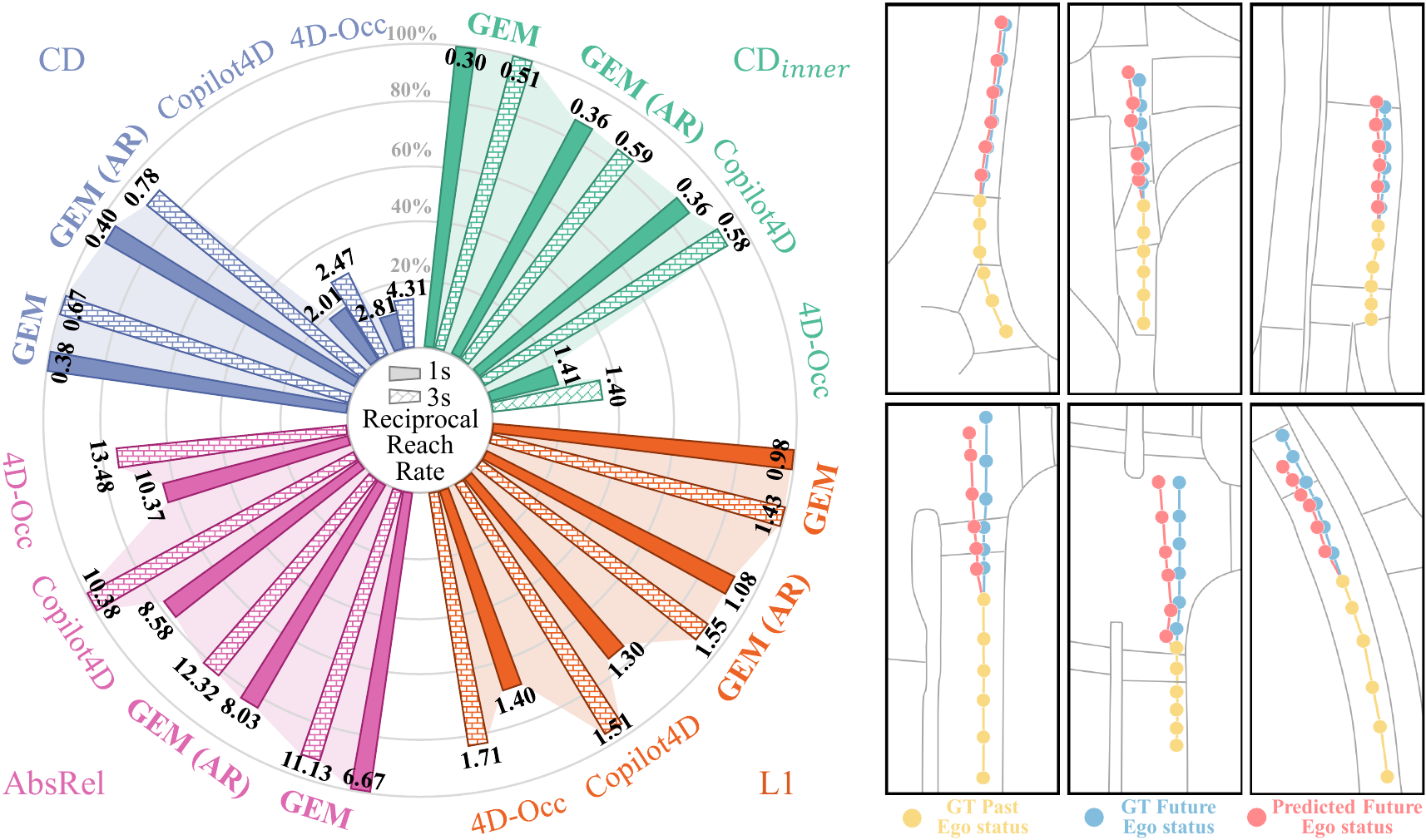}
\caption{The capability of \mtd{} to perform autonomous rollout (AR) and future prediction. GEM (AR) uses an integrated planner to provide future ego status, instead of ground truth.}
\label{fig: planner}
\vspace{-0.17cm}
\end{figure}
\begin{figure}[!t]
\includegraphics[width=0.478\textwidth]{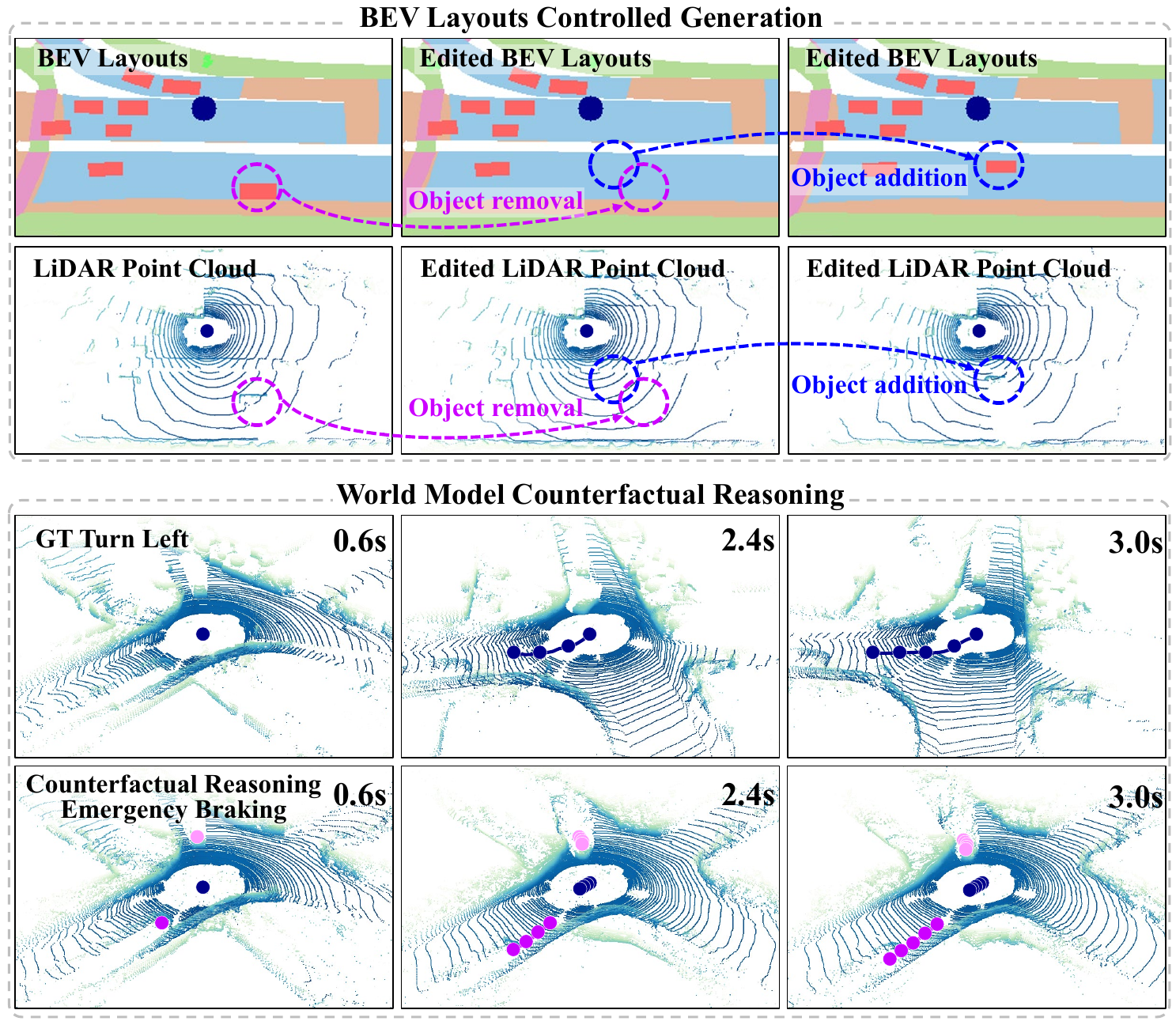}
\caption{The ability of \mtd{} to perform controllable generation and counterfactual reasoning, where royal blue dots represent the ego vehicle, and dots in other colors represent other vehicles.}
\label{fig: bev&reason}
\vspace{-0.17cm}
\end{figure}

This section shows autonomous rollout, BEV guided generation, and counterfactual reasoning of \mtd{}, assessing its ability to comprehend and model the world. 

The left panel of \cref{fig: planner} presents performance metrics normalized as percentages relative to the best result, highlighting variations across different approaches. 
Notably, our method equipped with autonomous ego poses planning shows only a slight performance drop, yet still surpasses other methods which rely on future ground truth inputs.
We also visualizes autonomous planning outcomes across diverse road conditions including straight roads, curves, and intersections on the right plane.  
These results collectively demonstrate that our method operates effectively in an autonomous rollout setting without requiring future ego-status information input, thereby offering greater scalability.

\cref{fig: bev&reason} demonstrates two capabilities of our method for controllable generation. 
The upper part shows BEV-layout-guided editing, allowing flexible object addition and removal to simulate intentionally controlled scenarios.
The lower part presents counterfactual predictions based on modified ego status, such as generating physically plausible interactions when replacing a left-turn with emergency braking.
Benefiting from coherent world understanding through explicit dynamic-static modeling, \mtd{} effectively generates ``what-if" scenarios, thereby paving the way for controlled synthesis of large-scale, scenario-specific data.

%% file: sec/5_conclu.tex
\section{Conclusion}
\label{sec:conclusion}
In this paper, we have proposed \mtd{}, a generative LiDAR world model that achieves high-accuracy future prediction with strong imaginative capacity. 
Within \mtd{}, a Mamba-based LiDAR tokenizer has been proposed to obtain LiDAR features efficiently while preserving the structural properties of LiDAR.
A dynamic-static separator has been further designed to obtain dynamic-static disentanglement. 
Guided by these cues, a tri-path deformable Mamba has been devised to perform structured driving world modeling.
Optionally, a planner and a BEV layout controller can be integrated in \mtd{} to support autonomous rollout and controllable generation.
Extensive experiments have demonstrated the superiority of \mtd{} in prediction accuracy, generation realism, and world understanding capabilities.